\documentclass[conference]{IEEEtran}
\usepackage{framed,multirow}
\usepackage{latexsym}
\usepackage{subfig}
\usepackage{graphicx}
\usepackage{epsfig}
\usepackage{amsmath}
\usepackage{amssymb}
\usepackage{float}
\usepackage{soul}
\usepackage{paralist}
\usepackage{array}
\usepackage{xcolor}
\usepackage{array}
\usepackage{mathtools}
\usepackage{bm}
\usepackage{xfrac}
\usepackage{float}

\ifCLASSINFOpdf
\else
\fi
\definecolor{commentcolor}{rgb}{0.2,0.2,1}

\newcolumntype{C}[1]{>{\centering\let\newline\\\arraybackslash\hspace{0pt}}m{#1}}

\graphicspath{ {./figures/}
				{./figures/samples/}}

\begin{document}
%
\title{Delineation of line patterns in \\images using \emph{B}-COSFIRE filters}

\author{\IEEEauthorblockN{Nicola Strisciuglio, Nicolai Petkov}
\IEEEauthorblockA{Johann Bernoulli Institute for Mathematics and Computer Science\\
University of Groningen, Netherlands\\
\emph{\{n.strisciuglio, n.petkov\}@rug.nl}}
}


%


\maketitle

\begin{abstract}
Delineation of line patterns in images is a basic step required in various applications such as blood vessel detection in medical images, segmentation of rivers or roads in aerial images, detection of cracks in walls or pavements, etc. In this paper we present trainable \textit{B}-COSFIRE filters, which are a model of some neurons in area V1 of the primary visual cortex, and apply it to the delineation of line patterns in different kinds of images.  \textit{B}-COSFIRE filters are trainable as their selectivity is determined in an automatic configuration process given a prototype pattern of interest. They are configurable to detect any preferred line structure (e.g. segments, corners, cross-overs, etc.), so usable for automatic data representation learning. 
We carried out experiments on two data sets, namely a line-network data set from INRIA and a data set of retinal fundus images named IOSTAR. The results that we achieved confirm the robustness of the proposed approach and its effectiveness in the delineation of line structures in different kinds of images.
\end{abstract}


%
\IEEEpeerreviewmaketitle

\section{Introduction}

The delineation of elongated patterns, such as line segments in images has applications in various fields. 
Line segments provide information about the geometric content of images and are considered important features for various applications. 
For instance, delineation algorithms are employed for the
detection and measure of cracks in materials~\cite{Mahadevan01} or in walls to estimate damages after earthquakes~\cite{walls}. Other applications involve automatic extraction of roads and rivers in aerial images for monitoring of road disruption~\cite{Mayer1998} or prevent flooding disasters~\cite{river}. In medical images, the delineation of blood vessels in retinal fundus or x-ray images serves as a basic step for further processing in automatic diagnostic systems.

A classical approach for extraction of lines and segments in images is the Hough transform, which maps the input image into a parameter space where lines of interest are detected~\cite{Duda72}. Other existing methods are based on filtering, region growing and mathematical morphology techniques, point and object processes and machine learning techniques.

Filtering techniques were based on multiscale analysis of local derivatives (Hessian matrix)~\cite{Frangi1998} or 2D-Gaussian kernels~\cite{HooverStare2000} to model the profile of line structures, with particular attention to blood vessels in retinal images. Multi-scale information about line width, size and orientation was also employed in region growing techniques~\cite{MartinezPerez}, while \emph{a-priori} information about the line network was combined with mathemathical morphology approaches in~\cite{ Mendonca2006}. Following the center-line of thick line structures was, instead, the basic idea of tracking methods~\cite{Chutatape1998}. 

Point (or object) processes were used for line and object detection although the simulation of their mathematical models is an expensive task, especially on large scenes. In~\cite{Lacoste2005}, line networks are modeled by an object process, where the objects correspond to interacting line segments. Extensions of the point processes were proposed in~\cite{Lafarge2010} and~\cite{Verdi2012} where a stochastic marked point process based on Gibbs model and a sampling procedure based on a Monte Carlo formalism were introduced, respectively. Point processes based on sampling junction-points in input images were combined with structural information provided by a graph-based representation~\cite{chai2013}. A graph-based method was also employed in automated reconstruction of tree structures using path classifiers and mixed integer programming~\cite{Turetken16}.

Machine learning techniques were employed in pixel-based approaches, where pixel-wise feature vectors were constructed and used in combination with classifier systems to discriminate between line and non-line pixels.
A $k$-NN classifier was used together with the responses of multiscale Gaussian filters and ridge detectors in~\cite{Niemeijer2004} and~\cite{StaalDrive2004}, respectively.  Multiscale Gabor wavelet coefficients were used as features to train a Bayesian classifier in~\cite{Soares2006}. An ensemble of bagged and boosted decision trees was proposed in~\cite{Fraz2012}. 
Recently, a deep learning classifier was trained with image patches of lines and used for the extraction of blood vessels from retinal fundus images~\cite{Liskowski2016}.

In this work, we present the \textit{B}-COSFIRE filters, originally proposed in~\cite{Azzopardi2015}, and apply them to the task of delineation of line structures in different kinds of images. The basic idea of \textit{B}-COSFIRE filters is inspired by the functions of some neurons in area V1 of the primary visual cortex, called \emph{simple cells}, devoted to detection of lines and bars of different thickness. The \textit{B}-COSFIRE filter is trainable as its structure is not fixed in the implementation, but it is rather learned in an automatic configuration process given a pattern of interest. 
The concept of trainable filters was previously introduced in~\cite{Azzopardi2013} and successfully employed in image processing~\cite{StrisciuglioVIP15,Guo2016}, object recognition~\cite{Robles2016} and adapted to audio analysis~\cite{StrisciuglioCOPE2016} applications.
Direct learning of the structure of filters from prototype patterns is a kind of representation learning, which allows to construct flexible methods for pattern recognition that can adapt to different applications. 

We demonstrate the effectiveness of \textit{B}-COSFIRE filters in the task of delineation of line structures in various types of images, such as retinal fundus, aerial, natural and indoor images. 
The results that we achieved, coupled with the small computational requirements, show the effectiveness of the \textit{B}-COSFIRE filters in the image delineation task and their usability in different applications.

The paper is organized as follows. In Section~\ref{sec:method}, we present the \textit{B}-COSFIRE filters while, in Section~\ref{sec:experiment}, we report and discuss the experimental results that we achieved on different types of images. Finally, we draw conclusions in Section~\ref{sec:conclusion}.

\section{Method}
\label{sec:method}

\subsection{Biological inspiration}
The characteristics of the \textit{B}-COSFIRE filters are inspired by functions of some neurons, called simple cells, in area V1 of the primary visual cortex~\cite{AzzopardiCORF2012}. Such neurons are known to be selective for elongated structures (lines, bars or contours) as described in the work of Hubel and Wiesel~\cite{Hubel1962}. 

\begin{figure}[!t]
	
	\footnotesize
	\setlength{\unitlength}{085mm}
	\includegraphics[width=\unitlength]{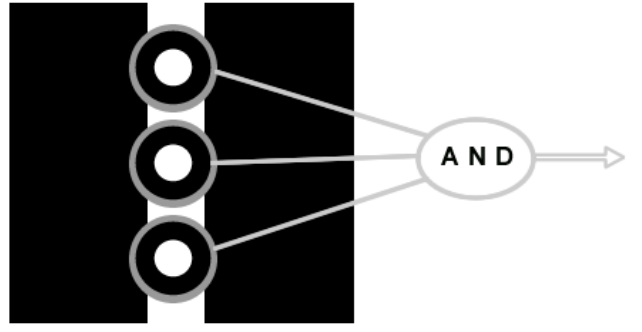}
   \caption{Sketch of a \textit{B}-COSFIRE filter. The responses of a group of DoG filters, shown in a gray disks, are taken along the bar. The outputs of such group (pool) are mutiplied to produce the output of the filter.    }
   \label{fig:rf}
\end{figure}

A \textit{B}-COSFIRE filter receives input from a pool of co-linearly aligned Difference of Gaussian filters, which are an accepted computational model of Lateral Geniculate Nucleus (LGN) cells~\cite{Rodieck1965} in the thalamus of the brain. Such cells detect contrast changes in the visual signal.  
In Fig.~\ref{fig:rf}, we show a sketch of the receptive field (RF) of a \textit{B}-COSFIRE filter in which each gray disk corresponds to a sub-unit that receives input from a center-on (or center-off) model LGN cell. The selectivity of a \textit{B}-COSFIRE filter is achieved by combining the responses of the sub-units aligned along the bar, as illustrated in Fig.~\ref{fig:rf}. 

The position of the sub-units in the model and their parameters are determined in an 
automatic configuration process in which an example bar of a given orientation and polarity is presented. 
This input stimulus determines a certain local configuration of model LGN cell activities in the RF of the concerned filter. The position of the considered LGN cell activities can be seen as the structure of the dendrites of simple cells. We create the model of a \textit{B}-COSFIRE filter by considering the spatial arrangement of this local configuration of sub-unit responses.
Finally, we compute the response of the considered \textit{B}-COSFIRE filter as the weighted geometric mean of the responses of its sub-units.

\subsection{\textit{B}-COSFIRE filters}
A \textit{B}-COSFIRE filter, originally proposed in~\cite{Azzopardi2015}, takes its input from the responses 
\begin{equation}
c_{\sigma}(x, y) \overset{\text{def}}{=} \lvert I \ast DoG_{\sigma}\lvert^{+}
\end{equation}
of a group of Difference-of-Gaussians $(DoG)$ filters at certain positions with respect to the center of its area of support, computed on the input image $I$. The notation $\lvert \cdot \lvert^+$ indicates a  half-wave rectification operation, also known as Rectified Linear Unit (ReLU).

The DoG filter with standard deviation $\sigma$ of the outer Gaussian function is formally defined as:
\begin{equation}
DoG^{+}_{\sigma} = \frac{1}{2\pi\sigma^2} e^{\left(-\frac{x^2+y^2}{2\sigma^2}\right)} - \frac{1}{2\pi(0.5\sigma)^2} e^{\left(-\frac{x^2+y^2}{2\pi(0.5\sigma)^2}\right)}.
\end{equation}
We set the standard deviation of the inner Gaussian function to $0.5\sigma$, following the results reported in electrophysiological studies~\cite{Irvine1993, Xu2002}. 

An automatic configuration process performed on a prototype pattern (a synthetic bar in our case) determines the positions at which we consider the DoG responses in the \emph{B}-COSFIRE filter model. The configuration process is explained in the following. A reference point is chosen as center of the support of the filter and the local maxima of the DoG responses along a number of concentric circles around such point are considered.
The result of the configuration is a set of $3-$tuples: ${S = \{(\sigma_i,\rho_i,\phi_i)\mid i=1,\dots,n\}}$, where $\sigma_i$ is the standard deviation of the outer $DoG$ function and $(\rho_i,\phi_i)$ are the polar coordinates of the $i$-th considered $DoG$ response with respect to the center of support of the filter. For further details about the configuration step, we refer the reader to~\cite{Azzopardi2015}.

We formally define the response of a \textit{B}-COSFIRE filter as the geometric mean of the responses of its sub-units:
\begin{equation}
r_{S}(x,y) \overset{\text{def}}{=} \left(\prod_{i=1}^{\mid S\mid}s_{\sigma_i,\rho_i,\phi_i}(x,y)\right)^{1/\mid S\mid}
\label{eq:response}
\end{equation}
where
\begin{equation}
\begin{split}
& \left. s_{\sigma_i,\rho_i,\phi_i}(x,y) =\right. \\ & \left. \operatorname*{max}_{x',y'}\{c_{\sigma_i}(x-\Delta x_i-x',y-\Delta y_i-y')G_{\sigma'}(x',y')\}, \right.
\end{split}
\label{eq:sub-units}
\end{equation}
with $-3\sigma' \leq x',  y' \leq 3\sigma'$, is the blurred and shifted response of the $i$-th sub-unit in the model $S$. The Gaussian weighting function $G(\cdot,\cdots)$ introduces tolerance in the position of the sub-units with respect to the ones configured in the model, accounting for robustness to deformations of the prototype pattern. The standard deviation $\sigma'$ of the function $G(x',y')$ is a linear function of the distance $\rho_i$ from the center of support of the filter: $\sigma' = \sigma_0' + \alpha \rho_i$.

The orientation selectivity of a \textit{B}-COSFIRE filter is determined by the orientation of the prototype pattern used for configuration. In order to achieve tolerance with respect to rotations of the pattern of interest, we manipulate the parameter $\phi_i$ in the model $S$ and obtain a new set ${R_{\psi}(S)=\{(\sigma_i,\rho_i,\phi_i + \psi)\mid i=1,\dots,n\}}$ with orientation preference $\psi$. We compute a rotation-tolerant response by taking the maximum response at every pixel among the responses of \textit{B}-COSFIRE filters with different orientation preferences:
\begin{equation}
\hat{r}_S(x,y) \overset{\text{def}}{=} \operatorname*{max}_{\psi \in \Psi}\big\{ r_{R_\psi(S)}(x,y) \big\}
\end{equation}
where $\Psi=\{0, \frac{\pi}{n_r}, \frac{2\pi}{n_r}, \dots, \frac{(n_r-1)\pi}{n_r}\}$ is a set of $n_r$ preferred orientations. In this work, we use the straightforward public Matlab implementation of a \textit{B}-COSFIRE filter\footnote{http://www.mathworks.com/matlabcentral/fileexchange/49172}.

\section{Experimental analysis}
\label{sec:experiment}

\subsection{Data}

We performed experiments on two data sets of images containing line networks, namely a data set distributed by INRIA\footnote{The images are available at the url http://www-sop.inria.fr/members/Florent.Lafarge/benchmark/evaluation.html} and a data set of retinal fundus images called IOSTAR~\cite{IOSTAR}.

The INRIA data set is composed of four images  that contain different types of line networks: the nerves of a leaf (Fig.~\ref{fig:leaf}), a tiled wall (Fig.~\ref{fig:tiles}) and two aerial images of a river and of a network of roads (Fig.~\ref{fig:river} and Fig.~\ref{fig:road}, respectively). Each image is provided with a ground truth image of the line network (see the images in the second column of Fig.~\ref{fig:samples}). An important contribution of the INRIA data set is that it allows to test the robustness of delineation algorithms on images from different fields and with diverse characteristics.

The IOSTAR data set contains 30 retinal fundus images with a resolution of $1024 \times 1024$ pixels. The images are acquired with an EasyScan camera, based on a Scanning Laser Ophthalmoscopy technique with a 45 degree Field of View (FOV). Each image is provided together with a ground truth image of the vessel tree and a mask of the retina field of view.

\subsection{Performance evaluation}

We threshold the response of the \textit{B}-COSFIRE filters to obtain a binary segmentation of the input images, in which pixels that belong to lines are separated from the ones that belong to the background.  
We compare the segmented output with the ground truth image and we count each pixel to belong to one of the following categories: true positive (TP), false positive (FP), true negative (TN) and false negative (FN).

In order to compare the performance of the proposed \textit{B}-COSFIRE algorithm with the one of other existing algorithms on the INRIA data set, we compute the true positive rate (TPR) and false positive rate (FPR):
\begin{equation}
TPR = \frac{TP}{TP+FN},~FPR = \frac{FP}{FP+TN} \nonumber
\end{equation}

For the evaluation of the performance of algorithms on the segmentation of blood vessels in retinal images, it is common to compute the accuracy (Acc), sensitivity (Se) and specificity (Sp) metrics, which are defined as: 
\begin{equation}
Acc = \frac{TP+TN}{N},~Se = \frac{TP}{TP+FN},~Sp = \frac{TN}{TN+FP}, \nonumber
\end{equation}

It is worth noting that in applications of delineation of elongated patterns from images, the number of line pixels is usually much lower than the number of the background pixels. The higher number of background pixels determines a bias in the evaluation of the performance results of delineation algorithms. Thus, as proposed in~\cite{Azzopardi2015}, we compute the Matthews correlation coefficient (MCC), which measures the performance of a binary classifier in the case the number of samples in the two classes is unbalanced. It is calculated as: 
\begin{equation}
MCC = \frac{TP / N - S \times P}{\sqrt{P \times S \times (1 - S) \times (1 - P)}}, \nonumber
\end{equation}
where $N = TN + TP + FN + FP$, $S =(TP+FN)/N$ and $P = (TP + FP) / N$.
We select the threshold for each image in the INRIA line-network data set as the one the maximize the value of the MCC measure. For the IOSTAR data set we choose a single threshold value for all the images in the data set as the one that maximize the average value of MCC.

\subsection{Results and discussion}

In Table~\ref{tab:results}, we report the results that we achieved on the images in the INRIA data set together with the required processing time. We also report the results and processing time achieved by other methods in the literature. The \textit{B}-COSFIRE filters achieved the highest MCC value on the leaf and tiles images, while it obtained comparable results with the ones of other approaches on the aerial images of rivers and roads. The time required by the \textit{B}-COSFIRE filter to process the images in the INRIA data set is much lower than the ones required by other methods, making it a suitable approach for large-scale applications.

\begin{table}[t]
  \renewcommand{\arraystretch}{1.5}
  \centering
\caption{Result comparison on the images of the INRIA line-network data set. The processing time required by existing algorithms is aso reported in seconds (s) and minutes (m).}
\begin{tabular}{c|l|ccc|c}
\multicolumn{1}{c}{~} & \bfseries Method & \bfseries TPR  & \bfseries FPR  & \bfseries MCC  & \bfseries Time  \\ \hline \hline
\multirow{3}{*}{\rotatebox{90}{\bfseries Leaf}} & \textit{B}-COSFIRE & $0.7599$ & $0.014$ & $\mathbf{0.7680}$ & $0.75$s \\
~ & Chai \emph{et al.}~\cite{chai2013} & $0.7060$ & $0.039$ & $0.6143$ & $73.5$s \\
~ & Verdie \emph{et al.}~\cite{Verdi2012} & $0.6490$ & $0.016$ & $0.68015$ & $32.7$s \\ \hline
\multirow{4}{*}{\rotatebox{90}{\bfseries Tiles}} & \textit{B}-COSFIRE & $0.8392$ & $0.0066$ & $\mathbf{0.8410}$ & $1$s \\
~ & Verdie \emph{et al.}~\cite{Verdi2012} & $0.783$ & $0.018$ & $0.7118$ & $103$s \\
~ & Chai \emph{et al.}~\cite{chai2013} & $0.649$ & $0.036$ & $0.5219$ & $227$s \\
~ & Lafarge \emph{et al.}~\cite{Lafarge2010} & $0.518$ & $0.046$ & $0.3902$ & $293$s \\ \hline
\multirow{5}{*}{\rotatebox{90}{\bfseries River}} & \textit{B}-COSFIRE & $0.5857$ & $0.0177$ & $0.4950$ &  $0.52$s \\
~ & Verdie \emph{et al.}~\cite{Verdi2012} & $0.7500$ & $0.024$ & $0.5542$ & $16.8$s \\
~ & Lafarge \emph{et al.}~\cite{Lafarge2010} & $0.5500$ & $0.015$ & $0.4981$ &  $108$s \\
~ & Lacoste \emph{et al.}~\cite{Lacoste2005} & $0.6500$ & $0.008$ & $\mathbf{0.641}$ &  $45$m \\
~ & Rochery \emph{et al.}~\cite{Rochery2006} & $0.4990$ & $0.01$ & $0.5073$ & $10$m \\ \hline
\multirow{5}{*}{\rotatebox{90}{\bfseries Road}} & \textit{B}-COSFIRE & $0.6275$ & $0.0082$ & $0.6433$ &  $4.4$s \\
~ & Verdie \emph{et al.}~\cite{Verdi2012} & $0.637$ & $0.004$ & $0.7412$ & $15.6$s \\
~ & Lafarge \emph{et al.}~\cite{Lafarge2010} & $0.658$ & $0.013$ & $0.665$ & $381$s \\
~ & Lacoste \emph{et al.}~\cite{Lacoste2005} & $0.812$ & $0.006$ & $\mathbf{0.8302}$ & $155$m \\
~ & Rochery \emph{et al.}~\cite{Rochery2006} & $0.49$ & $0.013$ & $0.5427$ & $60$m \\ \hline
\end{tabular}

\label{tab:results}
\end{table}

\begin{figure*}[!t]
	\centering
	\footnotesize
	\setlength{\unitlength}{42mm}
	\subfloat[]{\label{fig:leaf}
	\includegraphics[width=\unitlength]{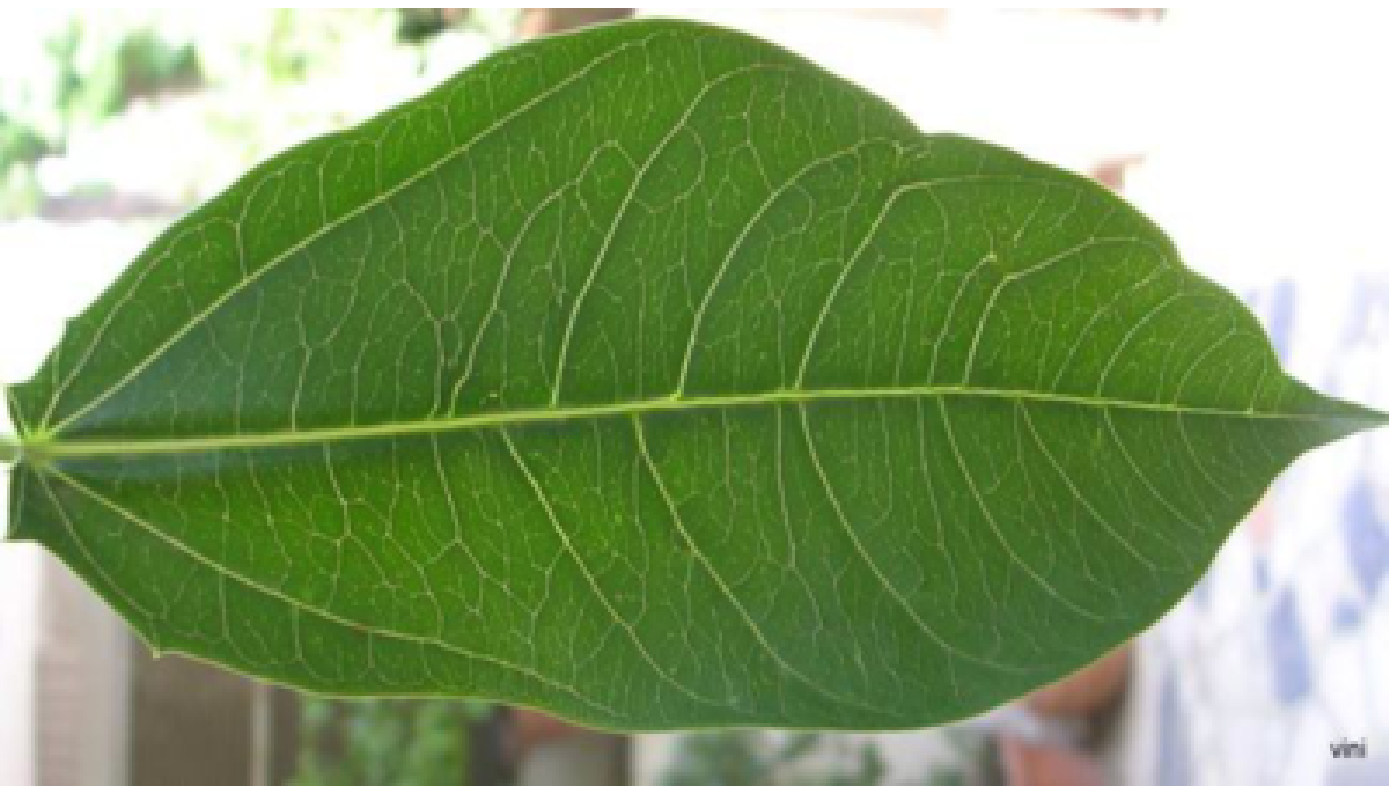}}
	~
	\subfloat[]{\label{fig:leaf-gt}
	\includegraphics[width=\unitlength]{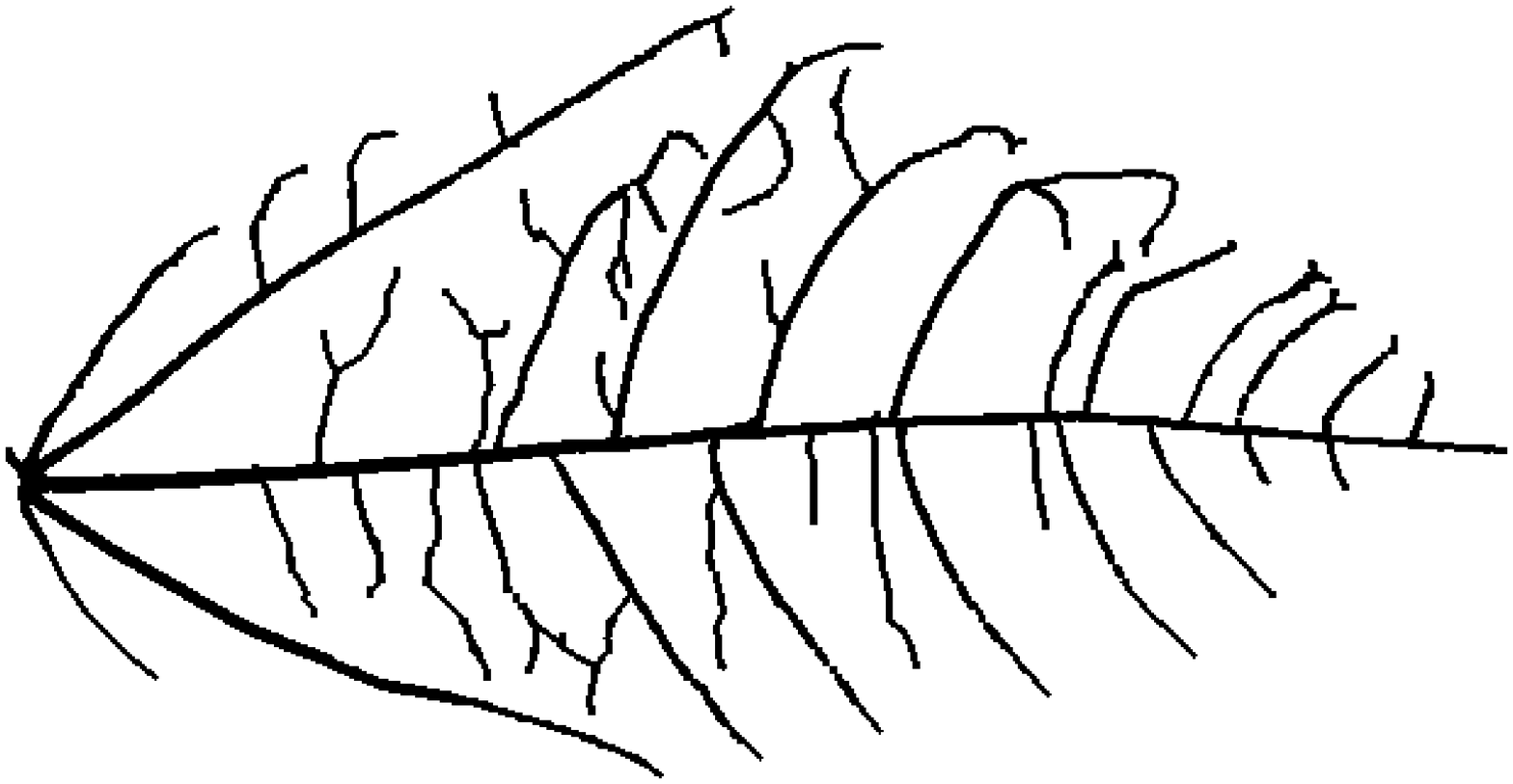}}
	~
	\subfloat[]{\label{fig:leaf-resp}
	\includegraphics[width=\unitlength]{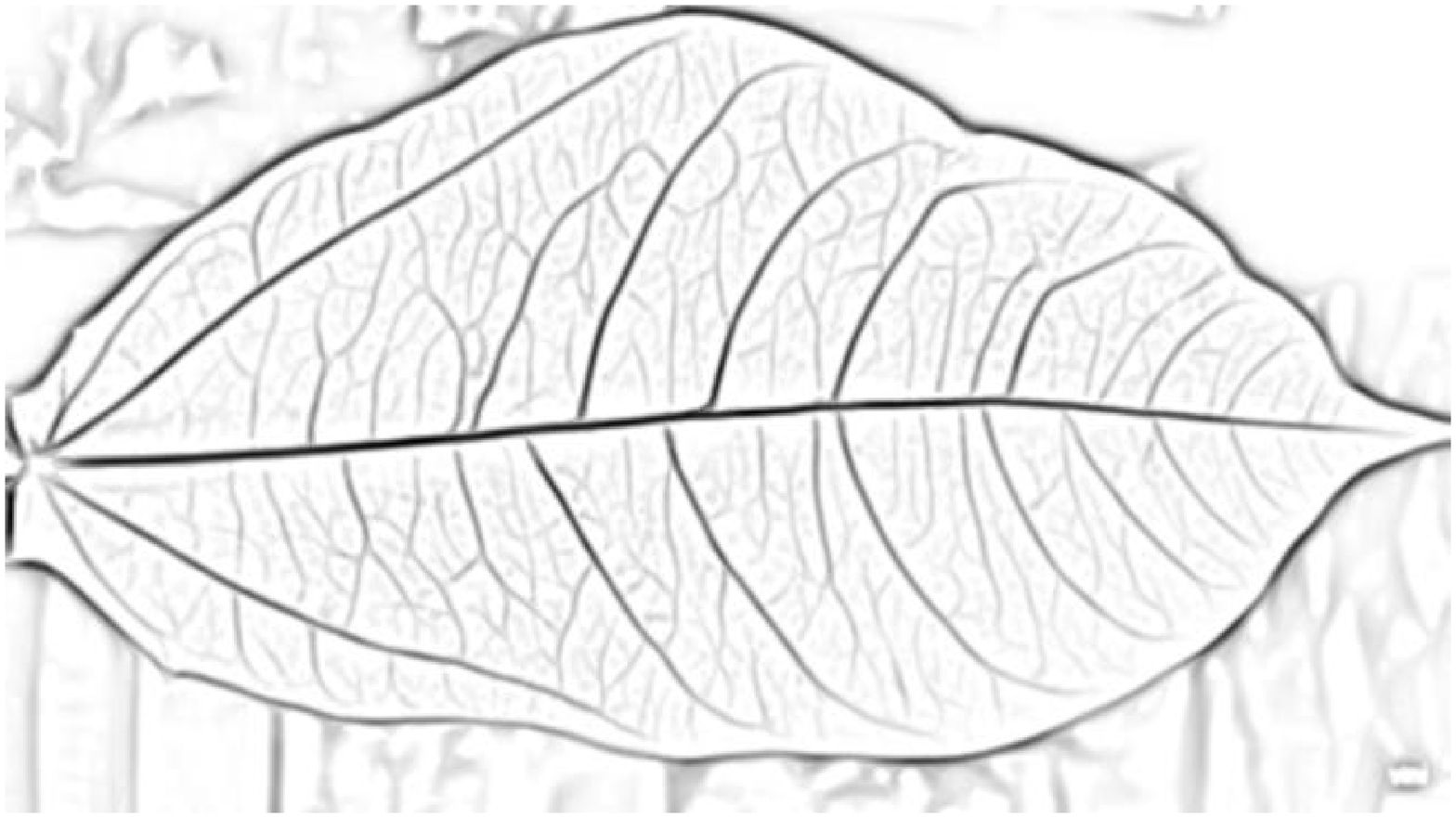}}
	~
	\subfloat[]{\label{fig:leaf-seg}
	\includegraphics[width=\unitlength]{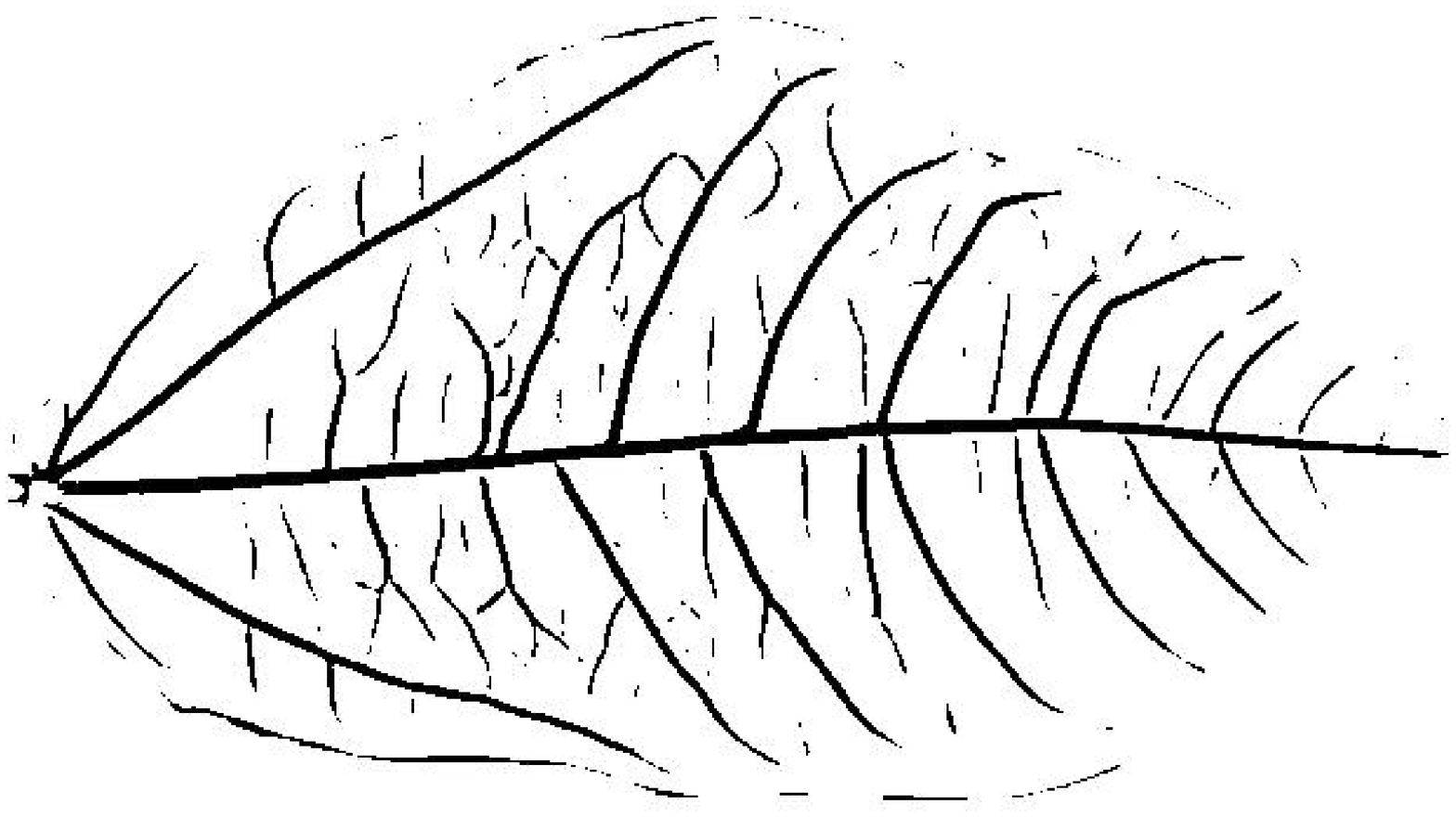}}
	
	\subfloat[]{\label{fig:tiles}
	\includegraphics[width=\unitlength]{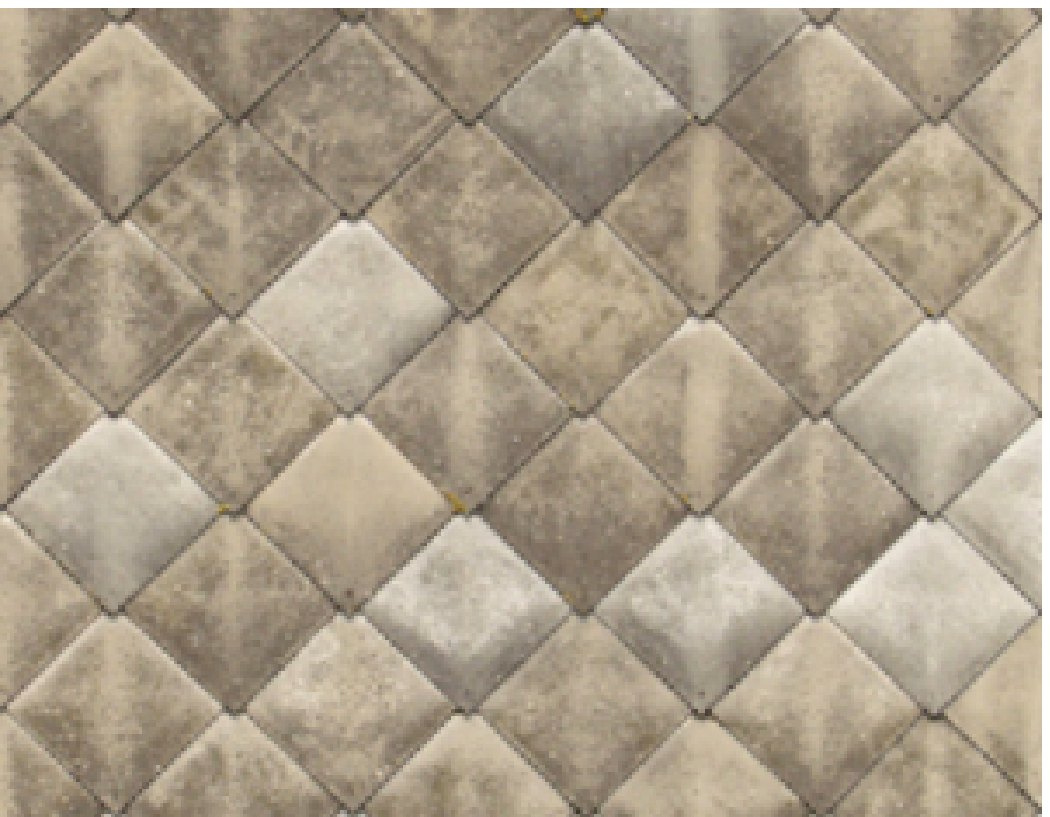}}
	~
	\subfloat[]{\label{fig:tiles-gt}
	\includegraphics[width=\unitlength]{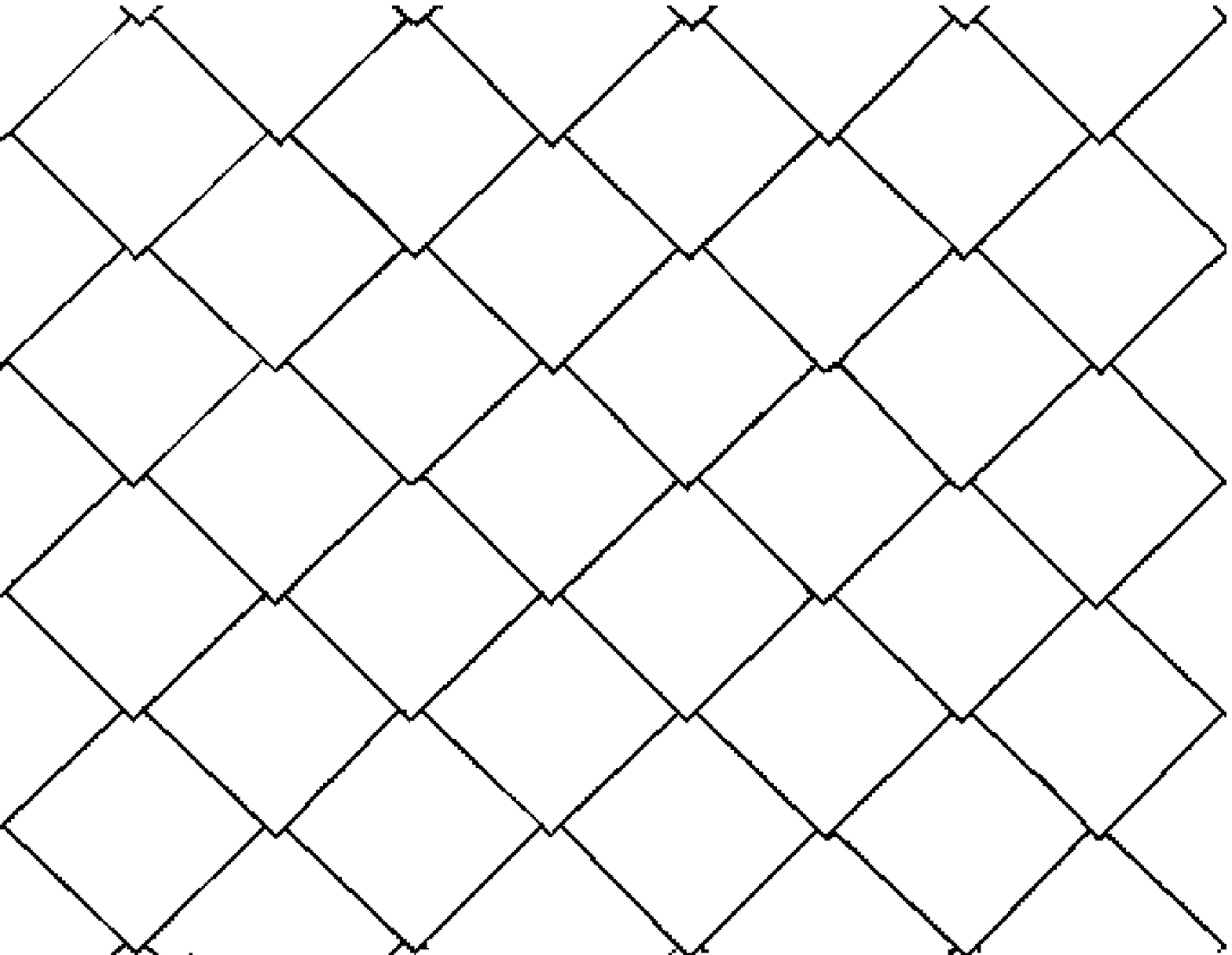}}
	~
	\subfloat[]{\label{fig:tiles-resp}
	\includegraphics[width=\unitlength]{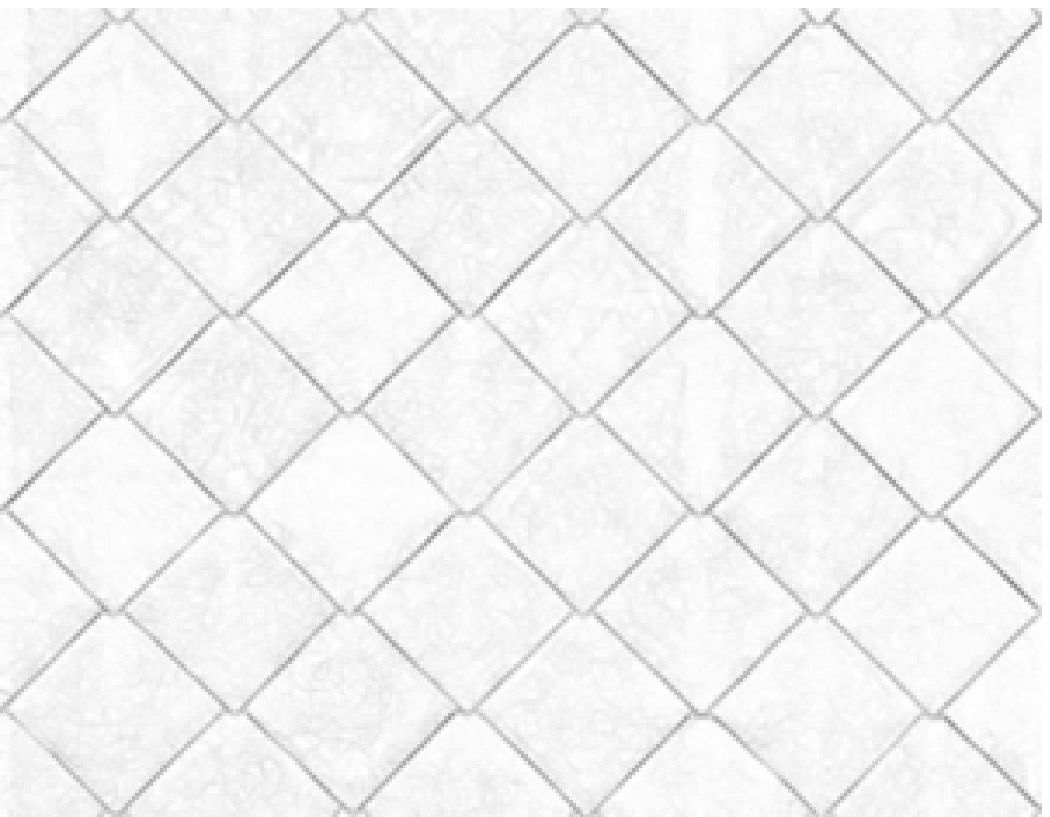}}
	~
	\subfloat[]{\label{fig:tiles-seg}
	\includegraphics[width=\unitlength]{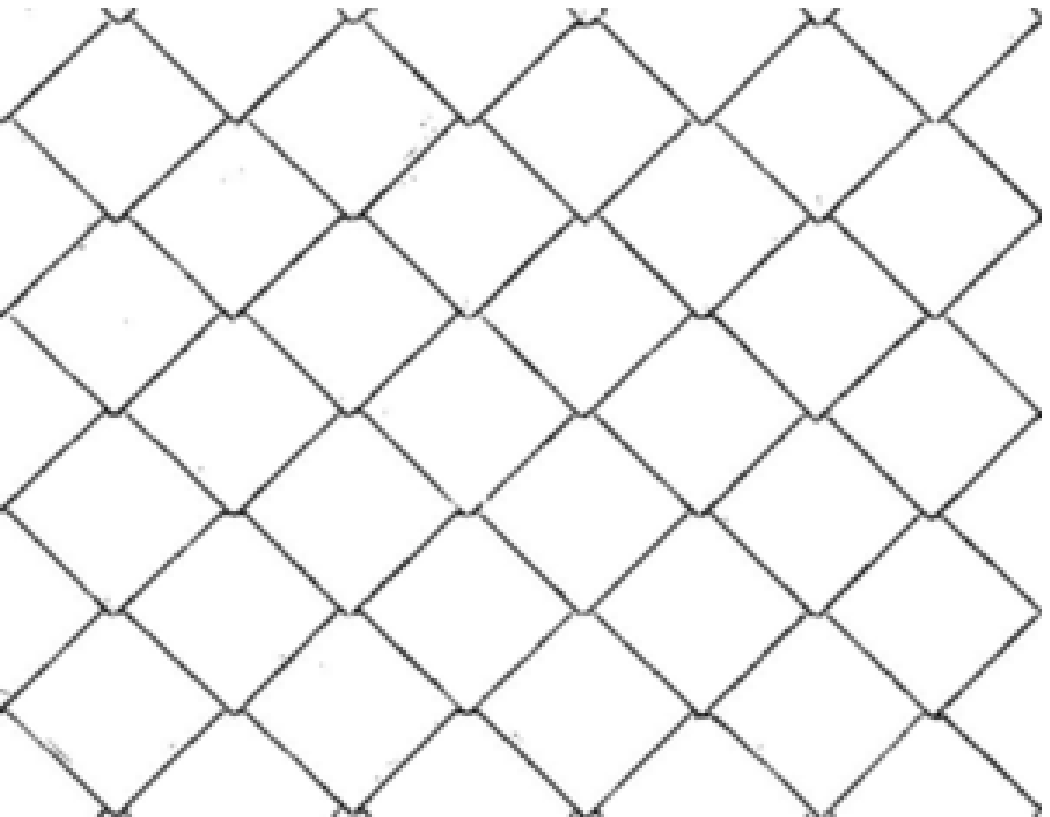}}	

	\subfloat[]{\label{fig:river}
	\includegraphics[width=\unitlength]{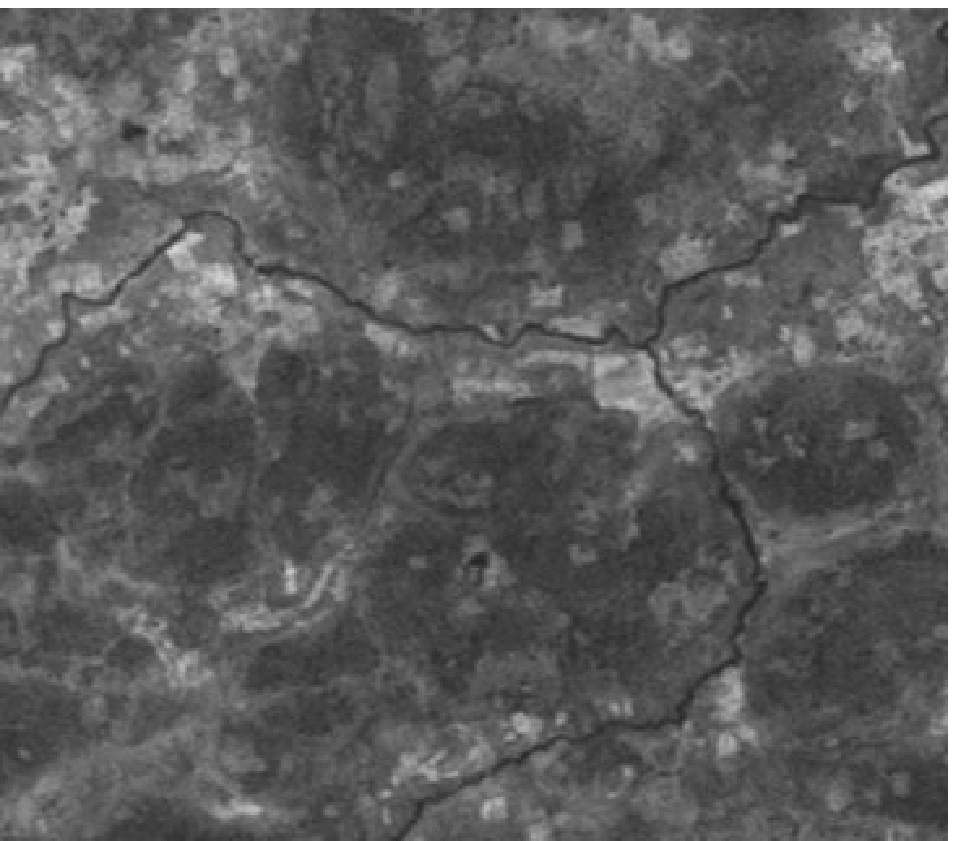}}
	~
	\subfloat[]{\label{fig:river-gt}
	\includegraphics[width=\unitlength]{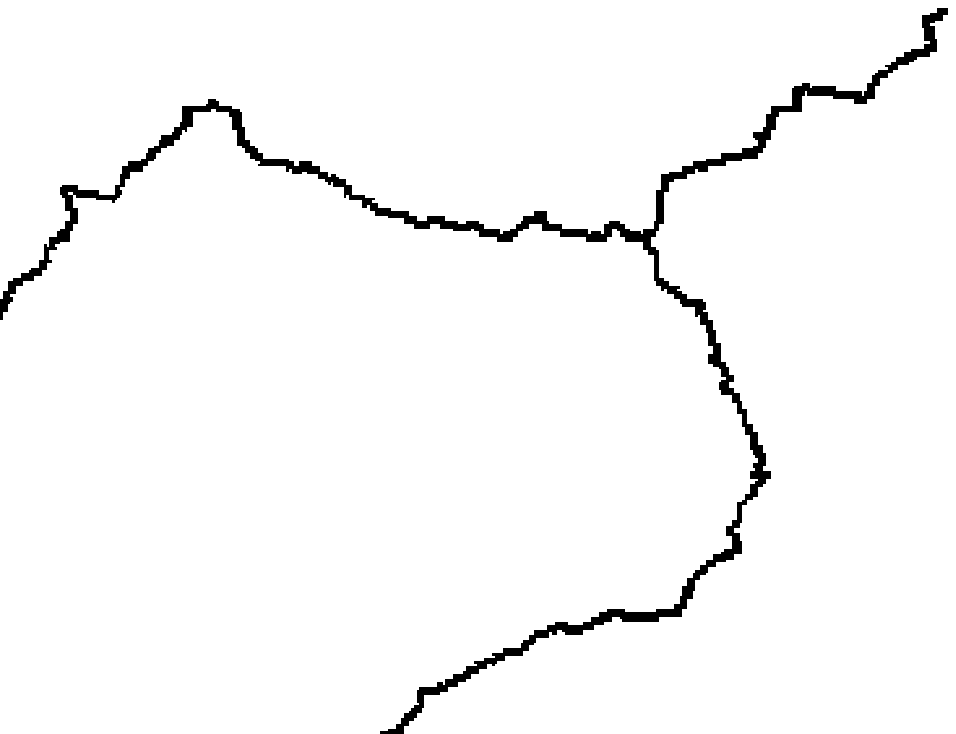}}
	~
	\subfloat[]{\label{fig:river-resp}
	\includegraphics[width=\unitlength]{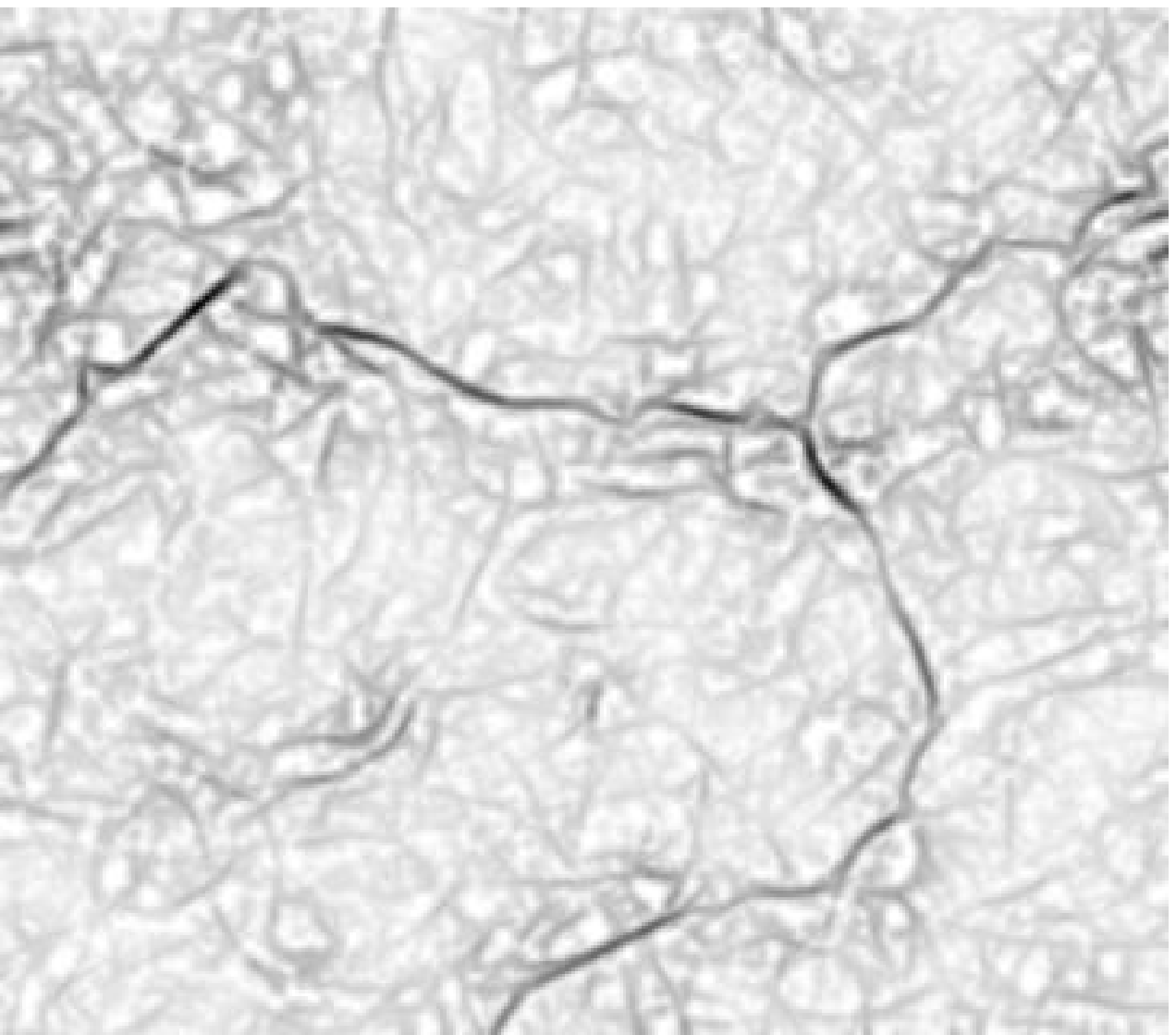}}
	~
	\subfloat[]{\label{fig:river-seg}
	\includegraphics[width=\unitlength]{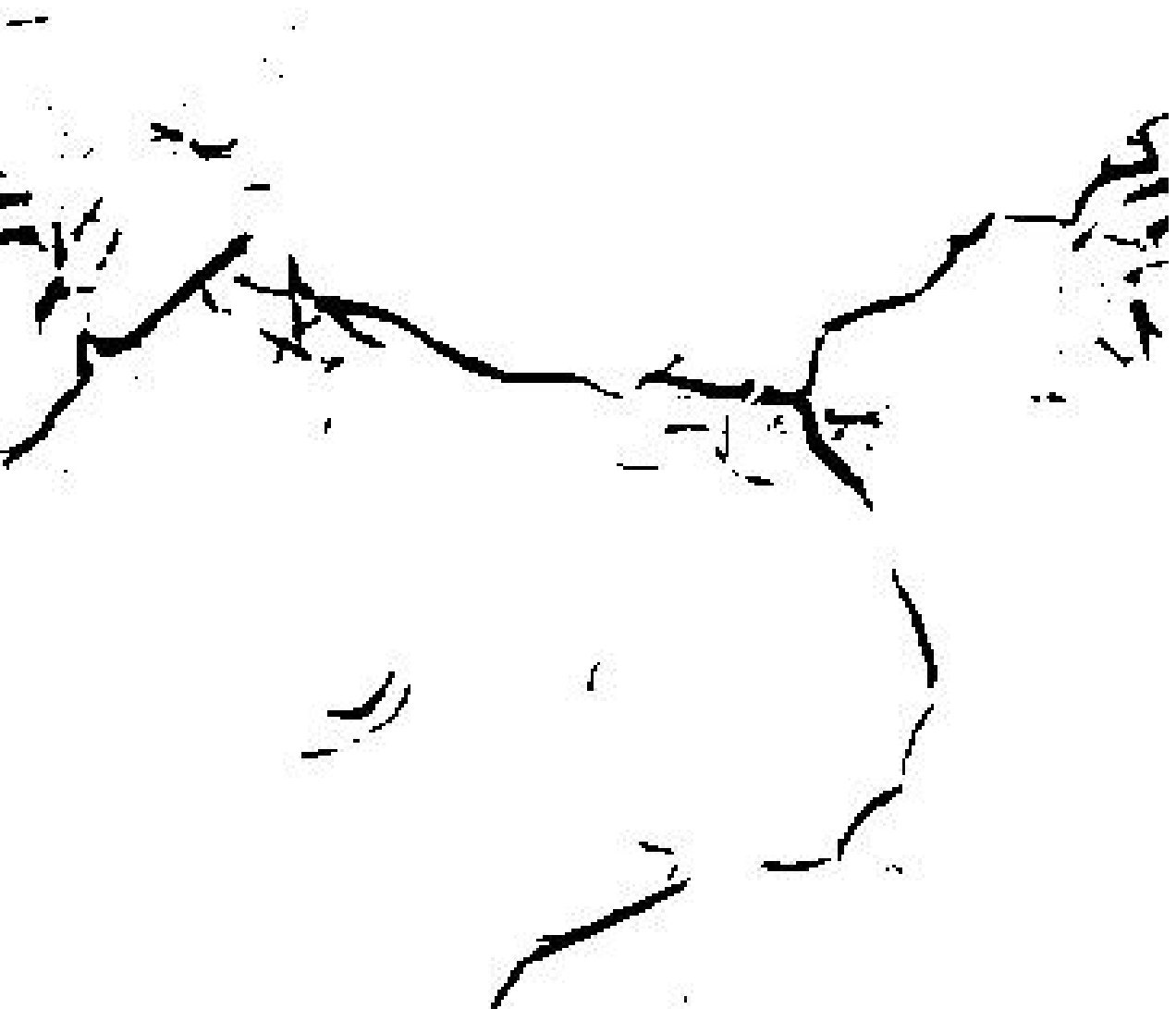}}
	
	\subfloat[]{\label{fig:road}
	\includegraphics[width=\unitlength]{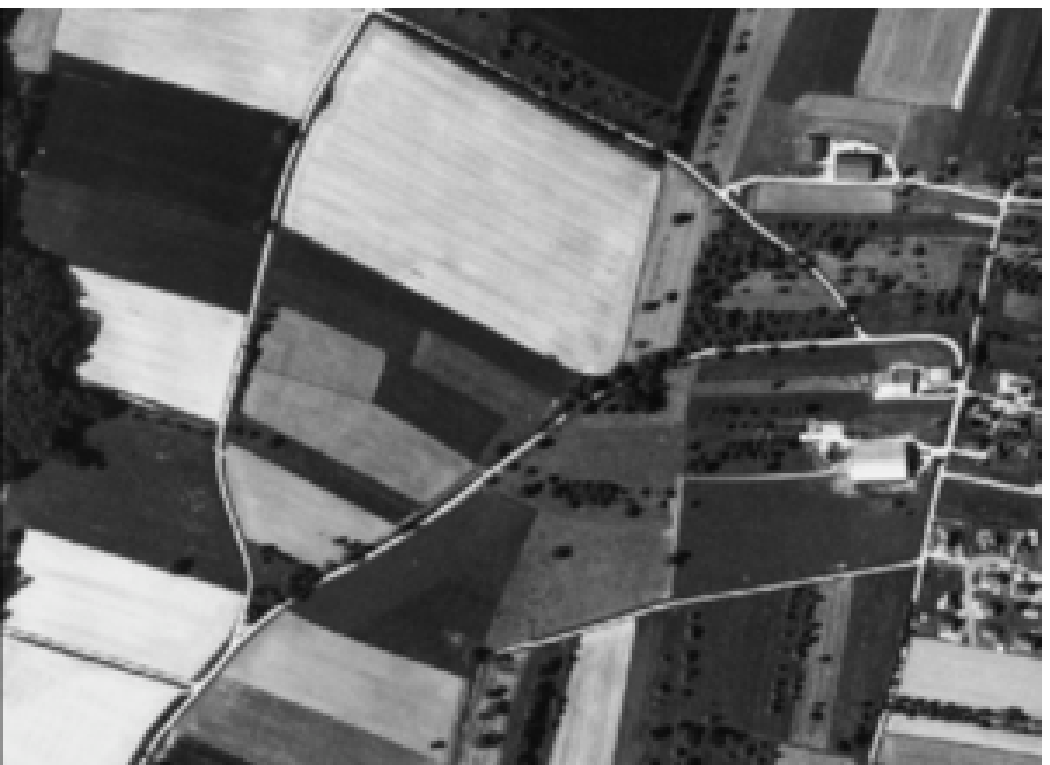}}
	~
	\subfloat[]{\label{fig:road-gt}
	\includegraphics[width=\unitlength]{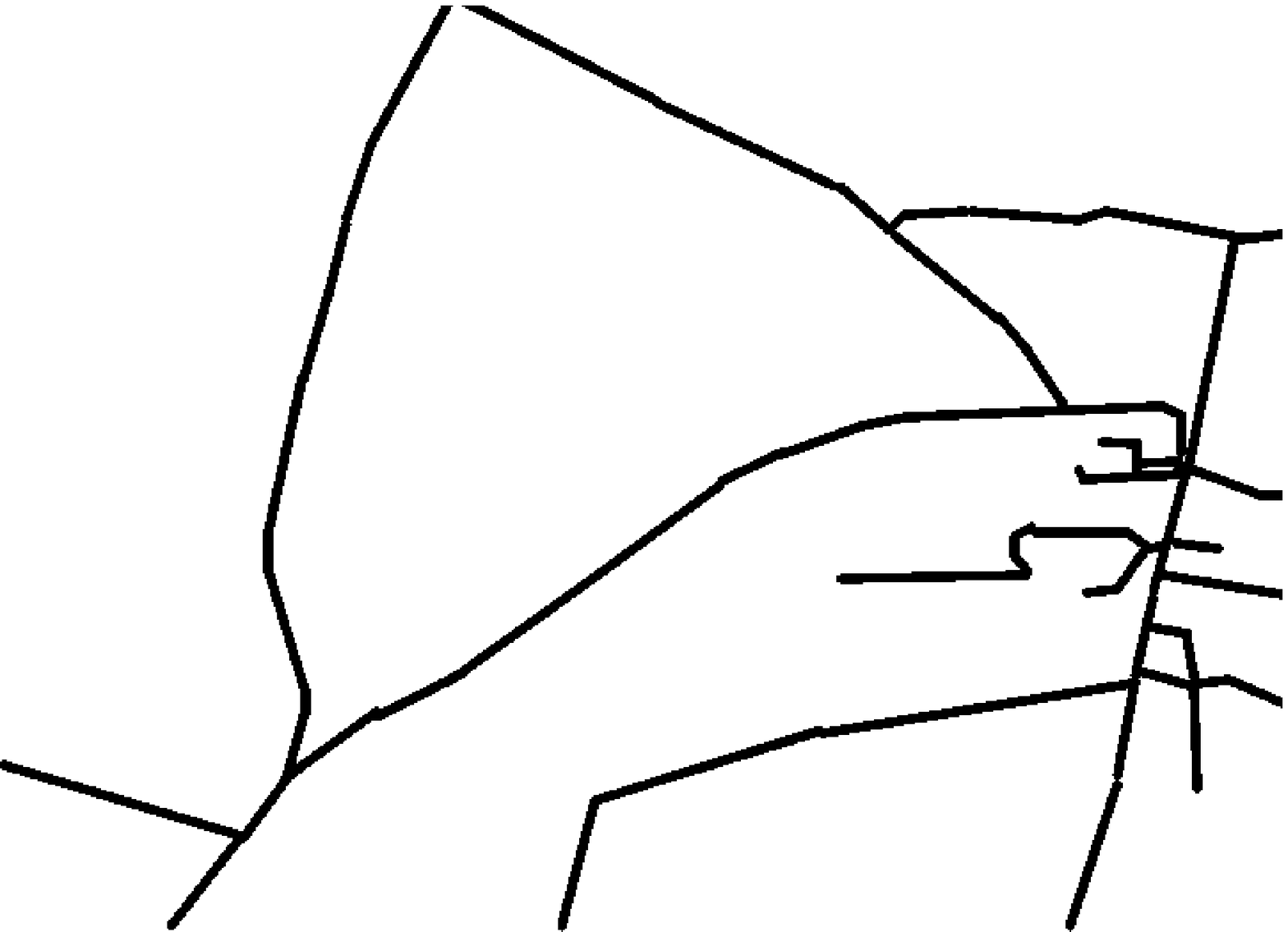}}
	~
	\subfloat[]{\label{fig:road-resp}
	\includegraphics[width=\unitlength]{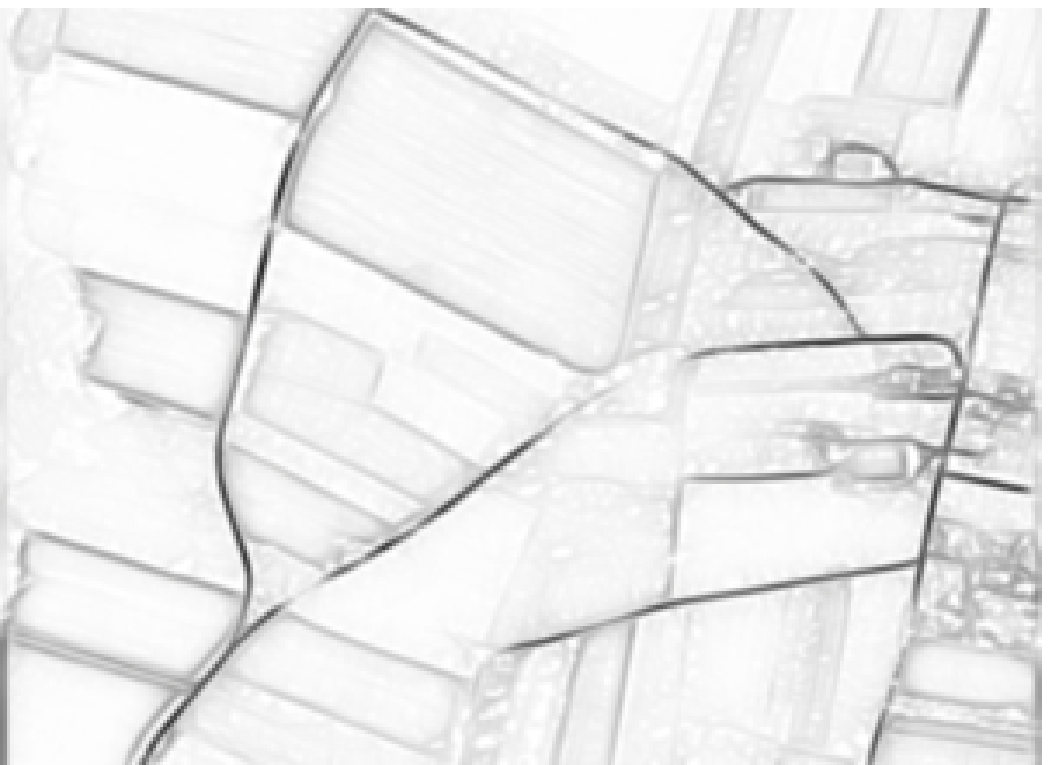}}
	~
	\subfloat[]{\label{fig:road-seg}
	\includegraphics[width=\unitlength]{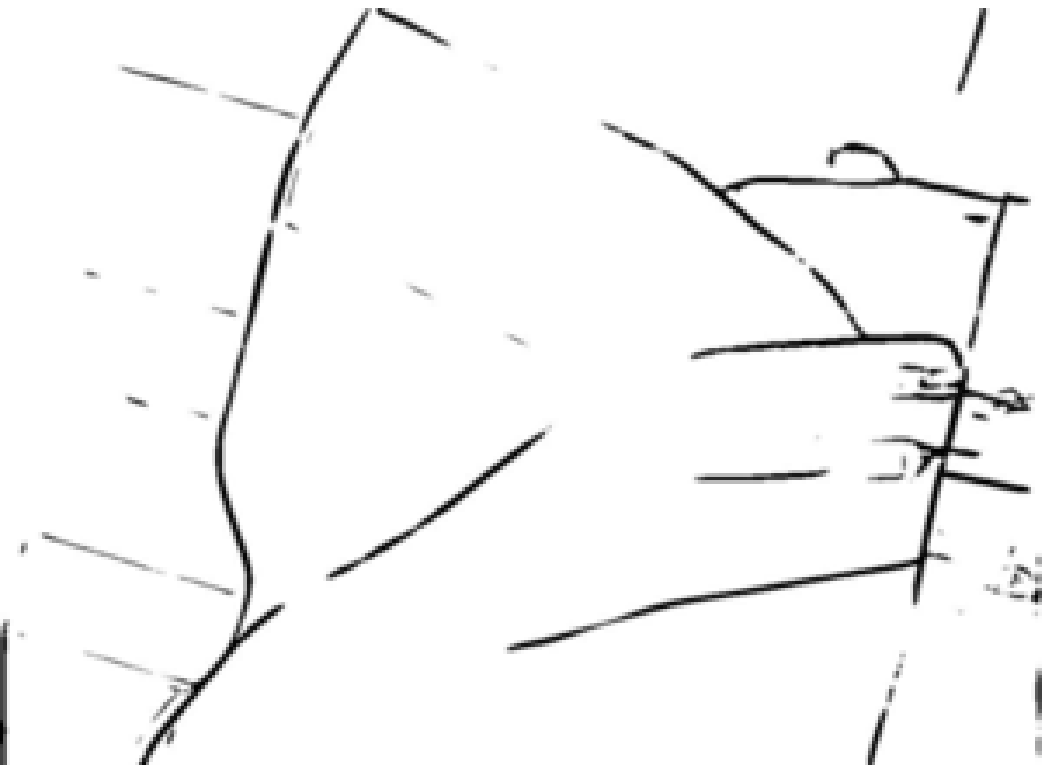}}
	
	\subfloat[]{\label{fig:retina}
	\includegraphics[width=\unitlength]{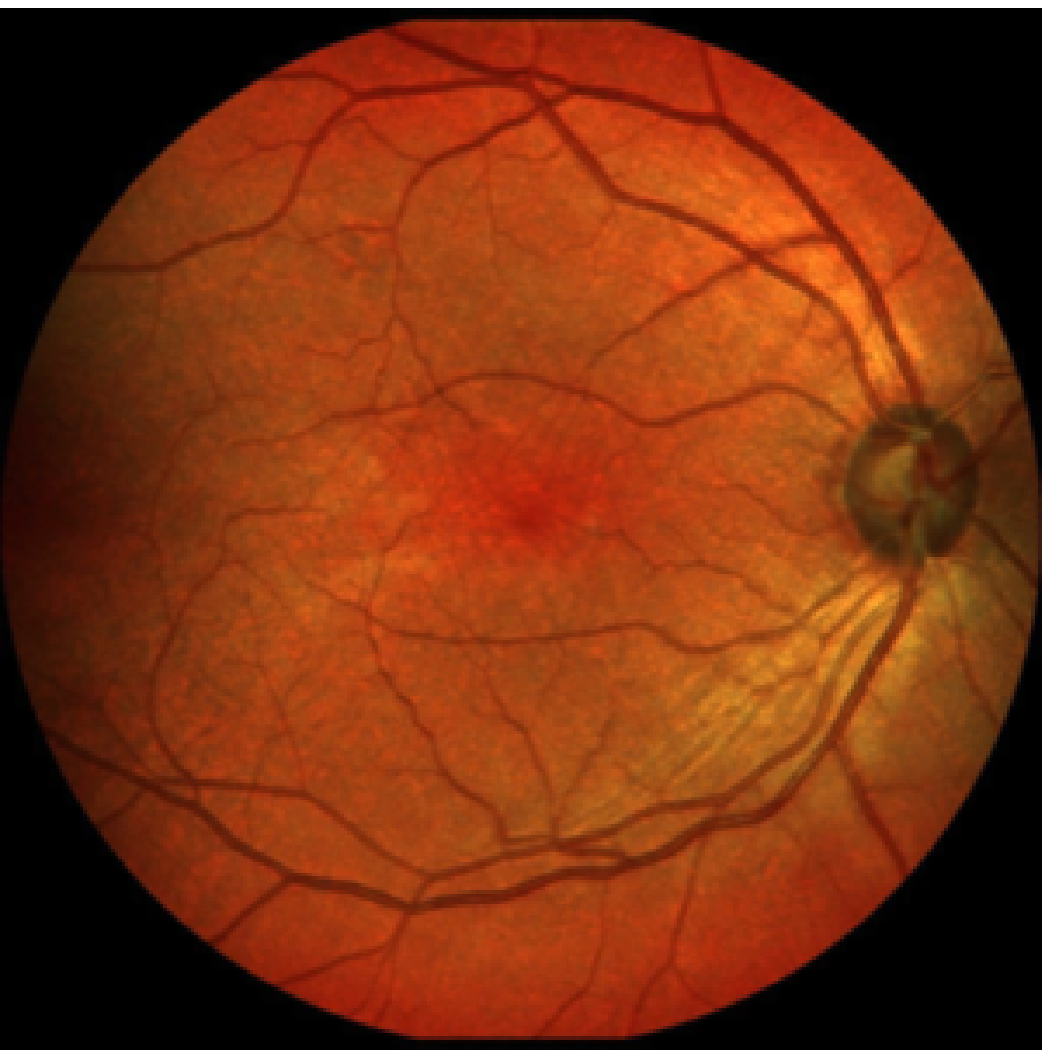}}
	~
	\subfloat[]{\label{fig:retina-gt}
	\includegraphics[width=\unitlength]{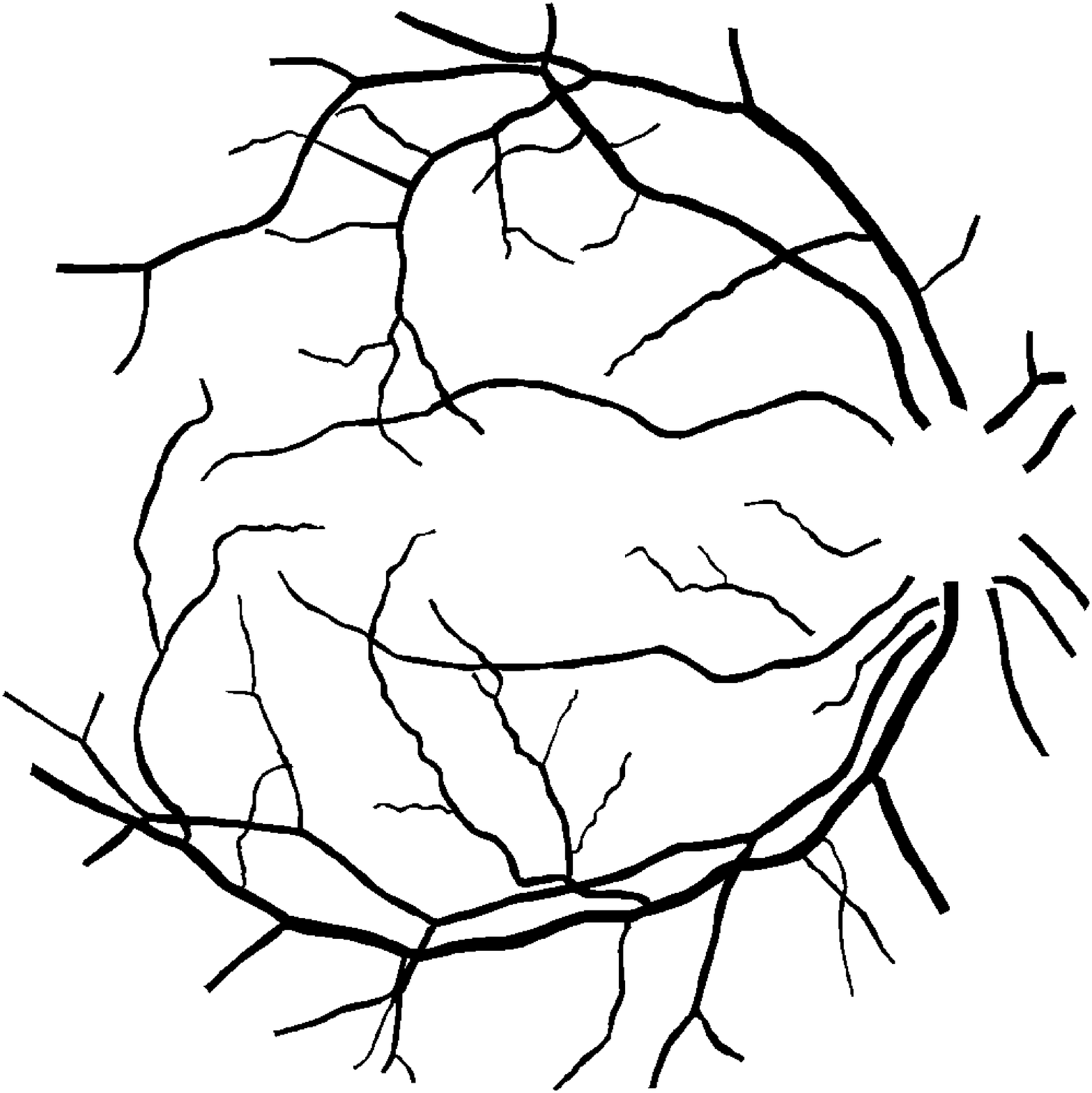}}
	~
	\subfloat[]{\label{fig:retina-resp}
	\includegraphics[width=\unitlength]{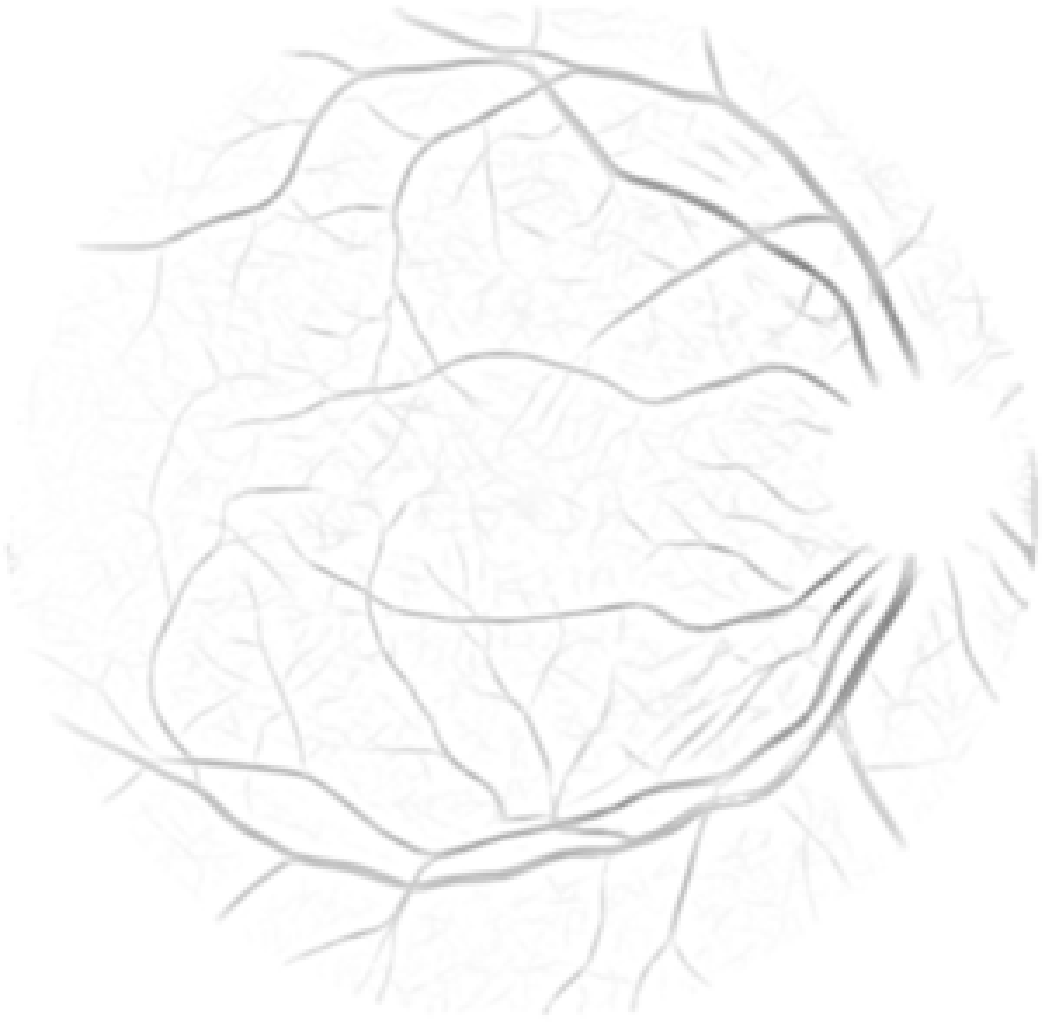}}
	~
	\subfloat[]{\label{fig:retina-seg}
	\includegraphics[width=\unitlength]{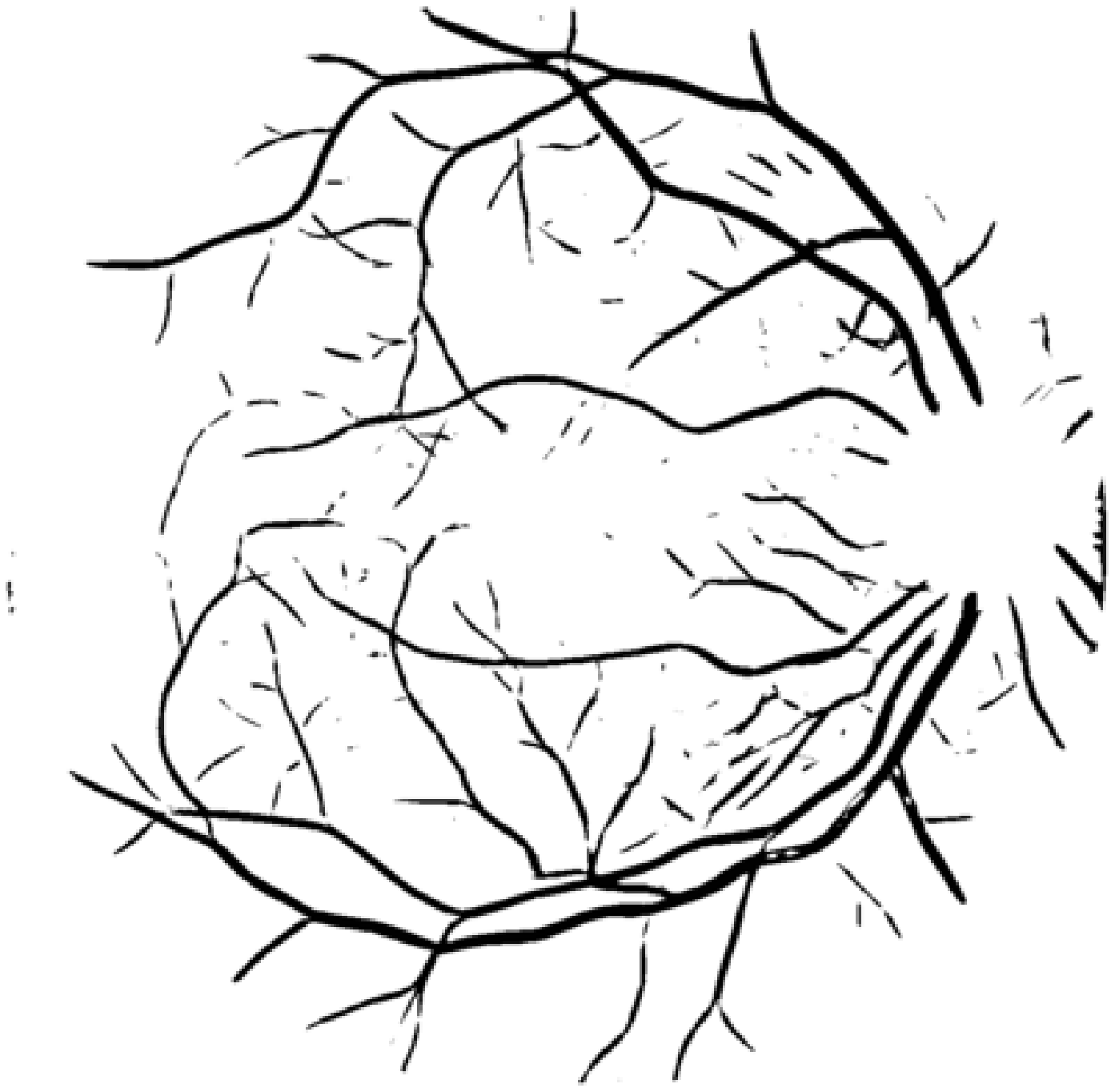}}
	
   \caption{Examples of different types of images used for the delineation experiments (first column) together with their manually segmented ground truth (second column). In the third column, we depict the responses of the \textit{B}-COSFIRE filters configured with the parameter values reported in Table~\ref{tab:params}, while in the fourth column we show the binary segmentation obtained by thresholding the \textit{B}-COSFIRE filter response.}
   \label{fig:samples}
\end{figure*}

In the third column of Fig.~\ref{fig:samples}, we show the responses of the \textit{B}-COSFIRE filters that are obtained by processing the images depicted in the first column. In the fourth column, instead we show the segmentation output obtained by thresholding the \textit{B}-COSFIRE filter response. We evaluate the performance of the proposed filters by comparing the segmented images with the ground truth images, reported in the second column of Fig.~\ref{fig:samples}. It is worth pointing out that the \textit{B}-COSFIRE filters that we configured in this work are selective for line patterns of given lengths and thickness. As it can be seen from the response images shown in Fig.~\ref{fig:road-resp} and~\ref{fig:river-resp}, the filters respond also to elongated linear  patterns that are not labeled as parts of interest in the ground truth. On one side, this corresponds to a decrease of performance for the specific application. On the other side, it shows the ability of the proposed filter to effectively delineate different kinds of elongated patterns. The lower MCC value achieved on the aerial images of rivers and roads w.r.t. other approaches is due to a missed reconstruction of occluded lines. The use of geometric mean, indeed, contributes to a low output response of the filter in case one or more expected sub-unit responses are missing. In such case, the effects of other combination functions, such as arithmetic mean, could be explored.

We report in Table~\ref{tab:retinaresults} the results achieved by the \textit{B}-COSFIRE filter on the IOSTAR data set in comparison with the ones achieved by the method published in~\cite{IOSTAR}. The value of MCC achieved by thresholding the response of the \textit{B}-COSFIRE filters is lower than the one achieved by the method proposed in~\cite{IOSTAR}. The lower performance of the proposed filters is mainly due to the characteristics of the images in the IOSTAR data set, which contain vessels with large range of thickness. A single \textit{B}-COSFIRE filter is able to detect elongated patterns with a limited range of thickness, around the one specified in the configuration step. In order to improve the delineation results, one can configure \textit{B}-COSFIRE filters with different values of the parameter $\sigma$, which control the selectivity for lines of a given thickness, and combine their responses in a multi-scale approach. However, as it can be seen in Fig.~\ref{fig:retina-resp}, a single \textit{B}-COSFIRE filter is able to effectively delineate a substantial quantity of vessels also in the images of the IOSTAR data set.
We processed the images of the first column of Fig.~\ref{fig:samples} with specific \textit{B}-COSFIRE filters, whose parameters are configured to corresponds to average characteristics of the lines in the images. In Table~\ref{tab:params}, we report the values of the parameters that we configured for the images of the concerned data sets.


In general, in order to improve the delineation performance in cases where the lines of interest have different scales (i.e. different thickness), one can configure a bank of \textit{B}-COSFIRE filters with various sets of configuration parameters and, then, employ filter selection techniques to determine a subset of relevant filters for the application at hand~\cite{Strisciuglio15, Strisciuglio2016}. 
As an example, the images in the IOSTAR data set contains vessel of various thickness and the use a set of filters with selectivity for vessels at different scales can improve the quality of the segmented images.

The trainable character of the \textit{B}-COSFIRE filter allows to configure filters selective to any line pattern of interest by presenting prototype samples to an automatic configuration process. This possibility is a kind of \emph{representation learning}, which involves the construction of a suitable data representation learned directly form training samples. In the design of traditional pattern recognition systems, a set of suitable features has to be engineered to describe the salient characteristics of the problem at hand. This process requires domain knowledge and it is not easy to optimize. Similarly to deep learning methodologies, the COSFIRE approach avoids engineering of hand-crafted features but it is rather able to determine important features directly from prototype training patterns. The automatic learning of suitable data representations allows to construct flexible and adaptive pattern recognition systems.

Although the computation of the response of a \textit{B}-COSFIRE filter is already efficient in its Matlab implementation and requires a small processing time (see Table~\ref{tab:results}), it can be further improved by a parallel software implementation. The blurring and shifting operations for different pairs $(\sigma, \rho)$ can be simultaneously processed on different processors.

\begin{table}[t]
  \renewcommand{\arraystretch}{1.5}
  \centering
\caption{Comparison of the results achieved on the IOSTAR data set of retinal fundus images.}
\begin{tabular}{c|C{12mm}cC{12mm}C{12mm}}
\bfseries Method & \bfseries Se  & \bfseries Sp  & \bfseries Acc  & \bfseries MCC  \\ \hline \hline
\textit{B}-COSFIRE & $0.7008$ & $0.9736$ & $0.9458$ & $0.6945$ \\
Zhang \emph{et al.}~\cite{IOSTAR} & $0.7787$ & $0.9710$ & $0.9512$ & $\bm{0.7327}$ \\ \hline
\end{tabular}
\label{tab:retinaresults}
\end{table}

\begin{table}[t]
  \renewcommand{\arraystretch}{1.5}
  \centering
\caption{Configuration parameters of the \textit{B}-COSFIRE filter for the processing of the considered images.}
\begin{tabular}{C{12mm}|C{12mm}cC{12mm}C{12mm}}
~ & \multicolumn{4}{c}{\bfseries Parameters} \\ \cline{2-5}
\bfseries Images & \bfseries $\bm{\sigma}$  & $\bm{\rho}$  & $\bm{\sigma_0}$  & $\bm{\alpha}$  \\ \hline \hline
Leaf & $2.9$ & $\{0,2,\dots,10\}$ & $2$ & $0.8$ \\
Tiles & $1.4$ & $\{0,2,\dots,8\}$ & $2$ & $0.4$ \\
River & $2.4$ & $\{0,2,\dots,12\}$ & $3$ & $0.8$ \\
Road & $1.7$ & $\{0,2,\dots,22\}$ & $4$ & $1.1$ \\
IOSTAR & $4.6$ & $\{0,2,\dots,22\}$ & $1$ & $0.3$ \\ \hline
\end{tabular}
\label{tab:params}
\end{table}

\section{Conclusion}
\label{sec:conclusion}

In this work we presented trainable \textit{B}-COSFIRE filters  and applied them in the task of delineating line patterns various kinds of images. 
They are trainable as their structure is learned from prototype samples in an automatic configuration step, rather than fixed in the implementation. 

We evaluated the performance on different types of images, such as aerial, indoor, natural and medical images.
The performance results that we achieved  demonstrate the effectiveness of the proposed method, and are comparable with the ones obtained by methods that were specifically designed to solve a particular problem. The robustness of the \textit{B}-COSFIRE filter in different kinds of images is attributable to the tolerance introduced in its application phase. The filter is indeed able to detect the same pattern used for the configuration process and also deformed versions of it. This properties and the obtained results make the proposed \emph{B}-COSFIRE filters applicable to various problems in which the delineation of elongated patterns is required.
The segmentation results that we obtained are coupled with good computational efficiency.



\bibliographystyle{IEEEtran}
\bibliography{references}

%

\end{document}